\documentclass[10pt,twocolumn,letterpaper]{article}
\usepackage{cvpr}              
\usepackage{graphicx}
\usepackage{amsmath}
\usepackage{amssymb}
\usepackage{booktabs}
\usepackage[pagebackref,breaklinks,colorlinks]{hyperref}

\usepackage[capitalize]{cleveref}
\crefname{section}{Sec.}{Secs.}
\Crefname{section}{Section}{Sections}
\Crefname{table}{Table}{Tables}
\crefname{table}{Tab.}{Tabs.}

\def\cvprPaperID{*****} 
\def\confName{CVPR}
\def\confYear{2022}

\begin{document}

\title{DAMO-YOLO : A Report on Real-Time Object Detection Design }

\author{Xianzhe Xu\textsuperscript{*}, Yiqi Jiang\textsuperscript{*}, Weihua Chen\textsuperscript{*}, Yilun Huang\textsuperscript{*},Yuan Zhang\textsuperscript{*}, Xiuyu Sun\textsuperscript{\dag}\\
Alibaba Group\\
}

\twocolumn[{%
\renewcommand\twocolumn[1][]{#1}%
\maketitle
\begin{center}
    \centering
    \captionsetup{type=figure}
    \includegraphics[width=1.0\textwidth,height=6.5cm]{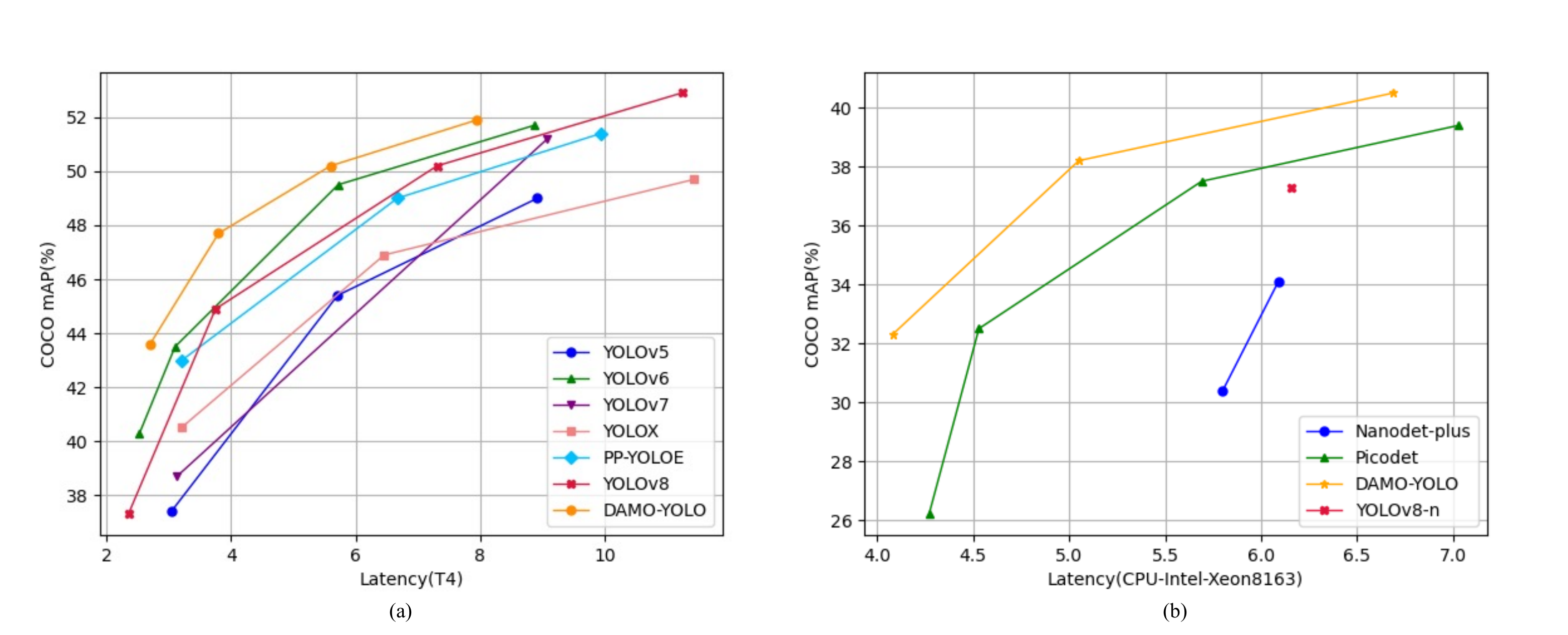}
    \captionof{figure}{Latency-accuracy trade-off of models for DAMO-YOLO and other state-of-the-art object detectors. (a) Comparision between the DAMO-YOLO-T/S/M/L and other SOTA general detectors on T4-GPU. (b) Comparision between the DAMO-YOLO-Ns/Nm/Nl and other SOTA light weight detectors on x86-CPU.  }
    \label{fig:coco_sota}
\end{center}%
}]
\def\thefootnote{*}\footnotetext{These authors contributed equally to this work}
\def\thefootnote{\dag}\footnotetext{Corresponding author \href{mailto:xiuyu.sxy@alibaba-inc.com}{xiuyu.sxy@alibaba-inc.com} }



\begin{abstract}

In this report, we present a fast and accurate object detection method dubbed \textbf{DAMO-YOLO}, which achieves higher performance than the state-of-the-art YOLO series. DAMO-YOLO is extended from YOLO with some new technologies, including Neural Architecture Search (NAS), efficient Reparameterized Generalized-FPN (RepGFPN), a lightweight head with AlignedOTA label assignment, and distillation enhancement.
In particular, we use MAE-NAS, a method guided by the principle of maximum entropy, to search our detection backbone under the constraints of low latency and high performance, producing ResNet-like / CSP-like structures with spatial pyramid pooling and focus modules.
In the design of necks and heads, we follow the rule of ``large neck, small head''.
We import Generalized-FPN with accelerated queen-fusion to build the detector neck and upgrade its CSPNet with efficient layer aggregation networks (ELAN) and reparameterization. Then we investigate how detector head size affects detection performance and find that a heavy neck with only one task projection layer would yield better results.
In addition, AlignedOTA is proposed to solve the misalignment problem in label assignment. And a distillation schema is introduced to improve performance to a higher level. 
Based on these new techs, we build a suite of models at various scales to meet the needs of different scenarios. For general industry requirements, we propose DAMO-YOLO-T/S/M/L. They can achieve 43.6/47.7/50.2/51.9 mAPs on COCO with the latency of 2.78/3.83/5.62/7.95 ms on T4 GPUs respectively. Additionally, for edge devices with limited computing power, we have also proposed DAMO-YOLO-Ns/Nm/Nl lightweight models. They can achieve 32.3/38.2/40.5 mAPs on COCO with the latency of 4.08/5.05/6.69 ms on X86-CPU. Our proposed general and lightweight models have outperformed other YOLO series models in their respective application scenarios. The code is available at \href{https://github.com/tinyvision/damo-yolo}{https://github.com/tinyvision/damo-yolo}.
\end{abstract}
\renewcommand{\thefootnote}{\arabic{footnote}} 
\section{Introduction}
\label{sec:intro}
Recently, researchers have developed object detection methods in huge progress \cite{girshick2015fast,ren2015faster,redmon2016yolo,liu2016ssd,yolov4,Wang_2021_CVPR,yolox}. While the industry pursues high-performance object detection methods with real-time constraints, researchers focus on designing one-stage detectors \cite{liu2016ssd,redmon2016yolo,lin2017focal,yolov3,yolov4}  with efficient network architectures ~\cite{detnas,spnas,maedet,NEURIPS2018_75fc093c,jiang2022giraffedet} and advanced training stages \cite{lin2017feature,tan2020efficientdet,jiang2022giraffedet,yolov3,ghiasi2019fpn}.
Especially, YOLOv5/6/7\cite{yolov5,yolov6,yolov7}, YOLOX \cite{yolox} and PP-YOLOE \cite{yoloe} have achieved significant AP-Latency trade-offs on COCO, making YOLO series object detection methods widely used in the industry.

Although object detection has achieved great progress, there are still new techs that can be brought in to further improve performance.
Firstly, the network structure plays a critical role in object detection. 
Darknet holds a dominant position in the early stages of YOLO history \cite{redmon2016yolo,yolov2,yolov3,yolov4,yolov5,yolox}. 
Recently, some works have investigated other efficient networks for their detectors, \ie, YOLOv6~\cite{yolov6} and YOLOv7~\cite{yolov7}. However, these networks are still manually designed. 
Thanks to the development of the Neural Architecture Search (NAS), 
there are many detection-friendly network structures found through the NAS techs~\cite{detnas,spnas,maedet}, which have shown great superiority over previous manually designed networks.
Therefore, we take advantage of the NAS techs and import MAE-NAS~\cite{maedet}\footnote{
\href{https://github.com/alibaba/lightweight-neural-architecture-search}{https://github.com/alibaba/lightweight-neural-architecture-search}. A demo can be found at the  \href{https://modelscope.cn/studios/damo/TinyNAS/summary}{ModelScope}.
} for our DAMO-YOLO. MAE-NAS is a heuristic and training-free neural architecture search method without supernet dependence and can be utilized to archive backbones at different scales. It can produce ResNet-like / CSP-like structures with spatial pyramid pooling and focus modules.
\begin{table}
    \begin{center}
    \caption{CSP-Darknet vs MAE-NAS Backbone under DAMO-YOLO framework with different scales.}
    \label{tab:head_ab_temp}
    \setlength{\tabcolsep}{3pt}
    \begin{tabular}{ccccccc}
    \toprule
     Scale & Depth&Backbone & AP & Latency(ms)    \\
    \midrule 
     S & 25&CSP-Darknet & 44.9 & 3.92 \\
    S & 25&MAE-Res & 45.6 & 3.83 \\
    S & 25&MAE-CSP & 45.3 & 3.79  \\
    M & 35&MAE-Res & 48.0  & 5.64   \\
    M & 35&MAE-CSP  & 48.7 & 5.60     \\
    \bottomrule
    \end{tabular}
    \end{center}
\end{table}

\begin{table}
    \begin{center}
    \caption{CSP-Darknet vs MAE-NAS Backbone under DAMO-YOLO framework with different scales. * denotes latency evaluated on X86-CPU.}
    \label{tab:backbone_ab}
    \setlength{\tabcolsep}{3pt}
    \begin{tabular}{ccccccc}
    \toprule
     Scale & Depth&Backbone & AP & Latency(ms)    \\
    \midrule 
     N  & 18 & MAE-Mob & 38.2 & 5.05\textsuperscript{*}\\
     N& 18 & MAE-Res & 37.4 & 5.89\textsuperscript{*} \\
     S & 25 & CSP-Darknet & 44.9 & 3.92 \\
    S & 25&MAE-Res & 45.6 & 3.83 \\
    S & 25&MAE-CSP & 45.3 & 3.79  \\
    M & 35&MAE-Res & 48.0  & 5.64   \\
    M & 35&MAE-CSP  & 48.7 & 5.60     \\
    \bottomrule
    \end{tabular}
    \end{center}
\end{table}

Secondly, it is crucial for a detector to learn sufficient fused information between high-level semantic and low-level spatial features, which makes the detector neck to be a vital part of the whole framework. The importance of neck has also been discussed in other works~\cite{jiang2022giraffedet,ghiasi2019fpn, wang2019panet, tan2020efficientdet}. Feature Pyramid Network (FPN)~\cite{ghiasi2019fpn} has been proved effective to fuse multi-scale features. Generalized-FPN (GFPN)~\cite{jiang2022giraffedet} improves FPN with a novel queen-fusion. In DAMO-YOLO, we design a Reparameterized Generalized-FPN (RepGFPN). It is based on GFPN but involved in an accelerated queen-fusion, the efficient layer aggregation networks (ELAN) and re-parameterization.

To strike the balance between latency and performance, we conducted a series of experiments to verify the importance of the neck and head of the detector and found that "large neck, small head" would lead to better performance.
Hence, we discard the detector head in previous YOLO series works~\cite{redmon2016yolo,yolov2,yolov3,yolov4,yolov5,yolox,yoloe}, but only left a task projection layer. The saved calculations are moved to the neck part.  
Besides the task projection module, there is no other training layer in the head, so we named our detector head as ZeroHead. Coupled with our RepGFPN, ZeroHead achieves state-of-the-art performance, which we believe would bring some insights to other researchers.

In addition, the dynamic label assignment, such as OTA~\cite{ge2021ota} and TOOD~\cite{tood}, is widely acclaimed and achieves significant improvement compared to the static label assignment~\cite{zhu2020autoassign}. However, the misalignment problem is still unsolved in these works. We propose a better solution called AlignOTA to balance the importance of classification and regression, which can partly solve the problem.

At last, Knowledge Distillation (KD) has been proved effective in boosting small models by the larger model supervision. This tech does exactly fit the design of real-time object detection. Nevertheless, applying KD on YOLO series sometimes can not achieve significant improvements as hyperparameters are hard to optimize and features carry too much noise. In our DAMO-YOLO, we first make distillation great again on models of all sizes, especially on small ones.


As shown in Fig.\ref{fig:coco_sota}, with the above improvements, we proposed a series of general and lightweight models that exceed the state of the arts by a large margin.

In summary, the contributions are three-fold:
\begin{enumerate} 
\item This paper proposes a new detector called \textbf{DAMO-YOLO}, which extends from YOLO but with more new techs, including MAE-NAS backbones, RepGFPN neck, ZeroHead, AlignedOTA and distillation enhancement.
\item DAMO-YOLO outperforms the state-of-the-art detectors (\eg YOLO series) on public COCO datasets in both general and lightweight categories.
\item A suite of models with various scales is presented in DAMO-YOLO (tiny/small/medium) to support different deployments. The code and pre-trained models are released at \href{https://github.com/tinyvision/damo-yolo}{https://github.com/tinyvision/damo-yolo}, with ONNX and TensorRT supported.
\end{enumerate} 

\begin{figure*}
    \begin{center}
    \includegraphics[width=1.0\textwidth]{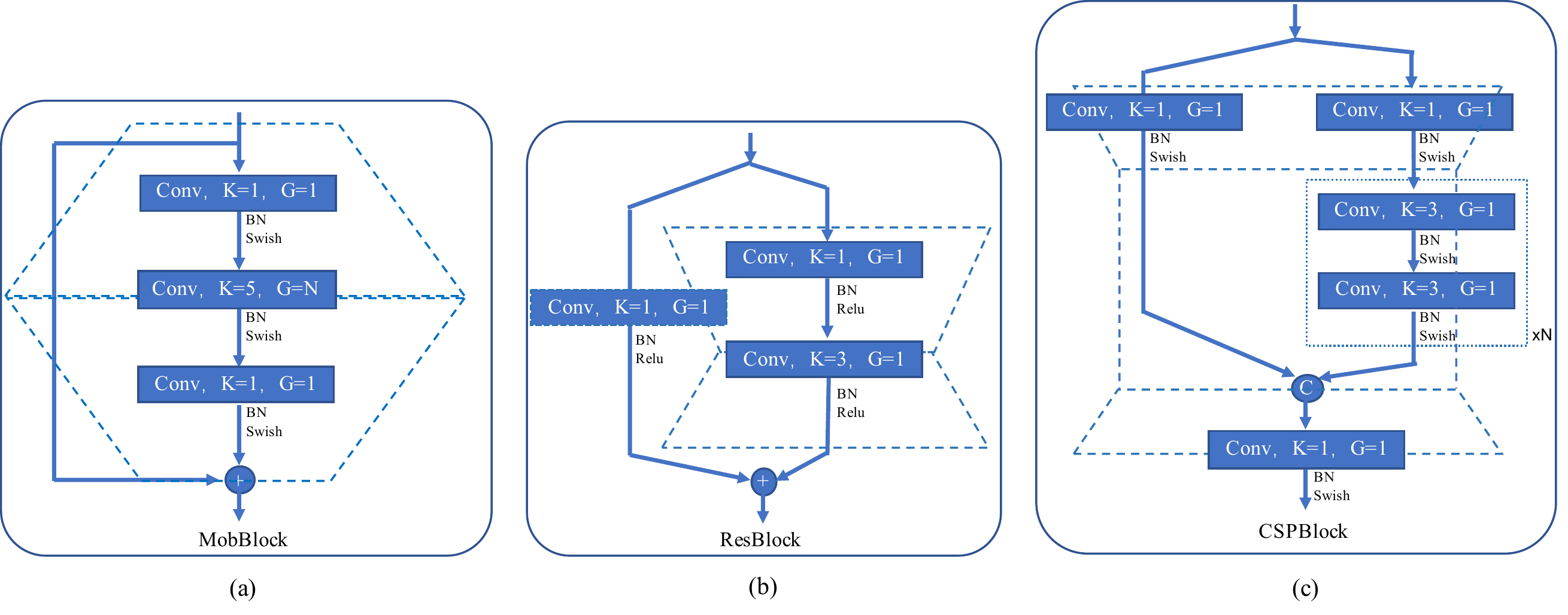} 
    \end{center}
       \caption{Different building block for MAE-NAS. (a) MobBlock is a variant of MobileNetV3 block and serves as the basic block for the lightweight model in DAMO-YOLO. (b) ResBlock is derived from ResNet. DAMO-YOLO-S and DAMO-YOLO-T is built upon it. (c) CSPBlock is derived from CSPNet, and due to its excellent performance in deep networks, it is used as a basic block in DAMO-YOLO-M and DAMO-YOLO-L models.}
    \label{fig:blocks}
\end{figure*}

\begin{figure*}
    \begin{center}
    \includegraphics[width=1.0\textwidth]{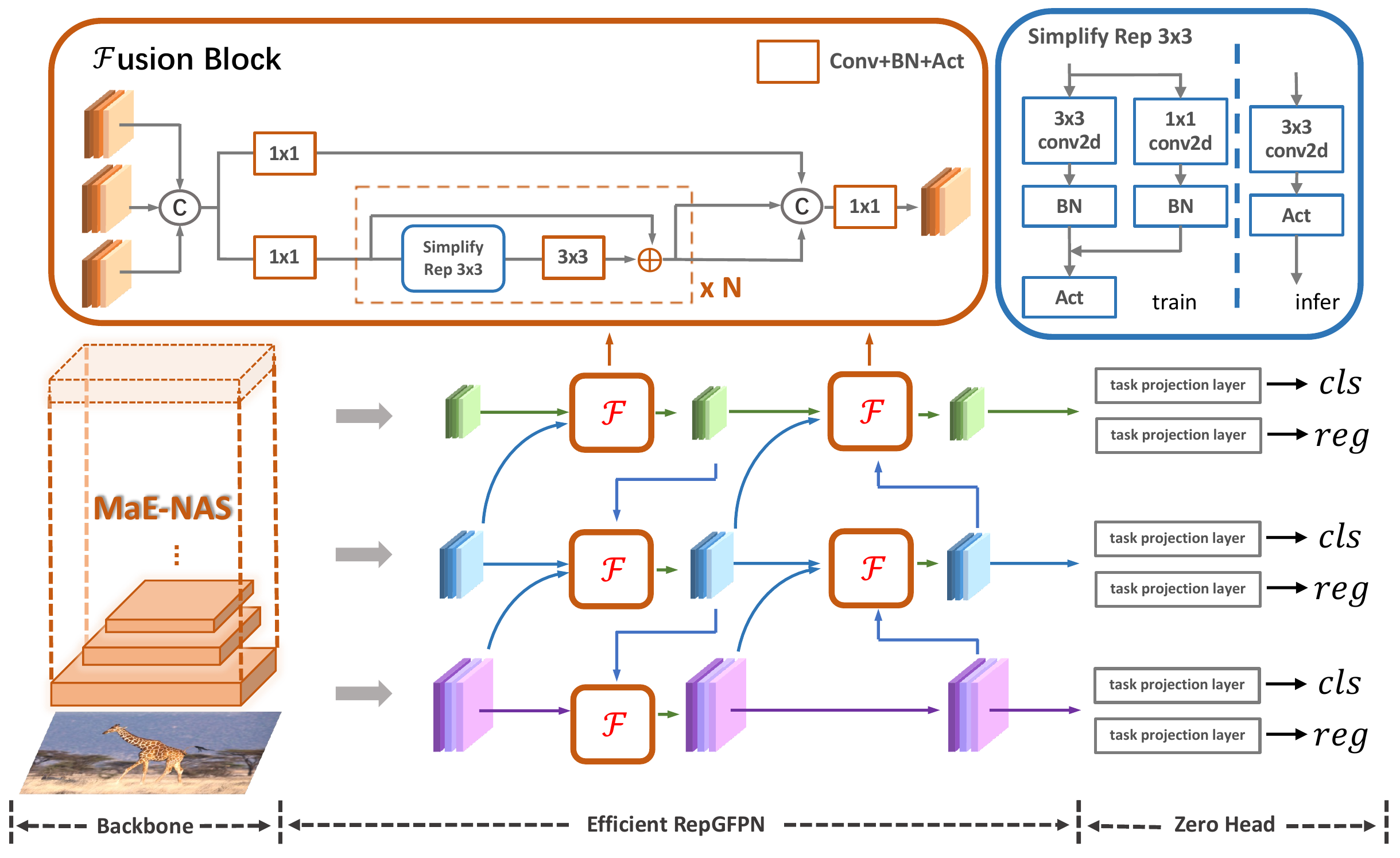} 
    \end{center}
    \vspace{-2mm}
    \caption{
       Overview of the network architecture of DAMO-YOLO. 1) MAE-NAS as backbone to extract multi-scale feature maps; 2) Efficient RepGFPN as neck to refine and fuse high-level semantic and low-level spatial features; 3) ZeroHead is presented which only contains a task projection layer for each loss.
       }
    \label{fig:network}
\end{figure*}
\section{DAMO-YOLO}
\label{sec:xxyolo}
In this section, we introduce each module of DAMO-YOLO in detail, including Neural Architecture Search (NAS) backbones, efficient Reparameterized Generalized-FPN (RepGFPN) neck, ZeroHead, AlignedOTA label assignment and distillation enhancement. The whole framework of DAMO-YOLO is displayed in Fig.\ref{fig:network}.

\subsection{MAE-NAS Backbone}
Previously, in real-time scenarios, designers relied on the Flops-mAP curve as a simple means of assessing model performance. However, the relationship between a model's flops and latency is not necessarily consistent. In order to enhance a model's real-world performance in industrial deployment, DAMO-YOLO prioritized the latency-MAP curve in the design process. 

Based on this design principle, we use MAE-NAS\cite{maedet} to obtain optimal networks under different latency budgets. MAE-NAS constructs an alternative proxy based on information theory to rank initialized networks without training. Therefore, the search process only takes a few hours, which is much lower than the training costs. Several basic search blocks are provided by MAE-NAS, such as Mob-block, Res-block and CSP-block, as shown in Fig.\ref{fig:blocks}. The Mob-block is a variant of MobileNetV3~\cite{howard2019searching} block, the Res-block is derived from ResNet~\cite{he2016deep}, and the CSP-block is derived from CSPNet~\cite{wang2020cspnet}.
The full supported block list can be found in MAE-NAS repository\footnote{\href{https://github.com/alibaba/lightweight-neural-architecture-search/blob/main/tinynas/models/README.md\#supported-blocks}{https://github.com/alibaba/lightweight-neural-architecture-search/blob/main/tinynas/models/README.md\#supported-blocks}}.


We found that applying different kinds of blocks in models of different scales achieves better trade-offs for real-time inference.
The performance comparisons among CSP-Darknet and our MAE-NAS backbones under our DAMO-YOLO with different scales are listed in Table.\ref{tab:head_ab_temp}.
In this table, ``MAE-Res'' means we apply Res-block in the MAE-NAS backbones, and ``MAE-CSP'' means we apply CSP-block in it. Besides, ``S'' (Small) and ``M'' (Medium) represent different scales of backbones. As shown in Table.\ref{tab:head_ab_temp}, the MAE-CSP obtained by MAE-NAS technology outperforms manually designed CSP-Darknet in terms of both speed and accuracy, demonstrating the superiority of MAE-NAS technolog. Moreover, we can observe from Table.\ref{tab:head_ab_temp} that using Res-block on smaller models can achieve a better trade-off between performance and speed than CSP-block, while using CSP-block can significantly outperform Res-block on larger and deeper networks. As a consequences, 
we use ``MAE-Res'' in ``T'' (Tiny) and ``S'' models and ``MAE-CSP'' in ``M'' and "L" models in the final setting. 

When dealing with scenarios that only have limited computing power or GPU is not available, it is crucial to have models that meet strict requirements for computation and speed. To address this issue, we have designed a series of light weight model using Mob-Block. Mob-block is derived from MobleNetV3~\cite{howard2019searching}, which can significantly decrease model computation and is friendly with CPU devices.

\subsection{Efficient RepGFPN} 
Feature pyramid network aims to aggregate different resolution features extracted from the backbone, which has been proven to be a critical and effective part of object detection~\cite{ghiasi2019fpn, wang2019panet, tan2020efficientdet}.
The conventional FPN~\cite{ghiasi2019fpn} introduces a top-down pathway to fuse multi-scale features. Considering the limitation of one-way information flow, PAFPN~\cite{wang2019panet} adds an additional bottom-up path aggregation network, but with higher computational costs. BiFPN~\cite{tan2020efficientdet} removes nodes that only have one input edge, and adds skip-link from the original input on the same level. 
In~\cite{jiang2022giraffedet}, Generalized-FPN (GFPN) is proposed to serve as neck and achieves SOTA performance, as it can sufficiently exchange high-level semantic information and low-level spatial information. 
In GFPN, multi-scale features are fused in both level features in previous and current layers. What's more, the $log_2(n)$ skip-layer connections provide more effective information transmission that can scale into deeper networks.
\begin{table}
    \begin{center}
    \caption{Ablation Study on the depth and width of our neck. ``Depth'' denotes the repeat times on the bottleneck of fusion block. ``Width'' indicates the channel dimensions of feature maps.}
    \label{tab:featuremap_scale}
    \setlength{\tabcolsep}{3pt}
    \begin{tabular}{c|ccccc}
    \toprule
    Depth & Width & Latency & FLOPs & AP    \\
    \midrule
    2 & (192, 192, 192) & 3.53 & 34.9 & 44.2    \\
    2 & (128, 256, 512) & 3.72 & 36.1 & 45.1    \\
    3 & (160, 160, 160) & 3.91 & 38.2 & 44.9     \\
    \textbf{3} & \textbf{(96, 192, 384)} & \textbf{3.83} & \textbf{37.8} & \textbf{45.6}  \\
    4 & (64, 128, 256) & 3.85 & 37.2 & 45.3     \\
    \bottomrule
    \end{tabular}
    \end{center}
\end{table}
When we directly replace modified-PANet with GFPN on modern YOLO-series models, we achieved higher precision. However, the latency of GFPN-based model is much higher than modified-PANet-based model. By analyzing the structure of GFPN, the reason can be attributed to the following aspects: 1) feature maps with different scales share the same dimension of channels; 2) the operation of queen-fusion can not meet the requirement for real-time detection model; 3) the convolution-based cross-scale feature fusion is not efficient.

Based on GFPN, we propose a novel Efficient-RepGFPN to meet the design of real-time object detection, which mainly consists of the following insights:
1) Due to the large difference in FLOPs from different scale feature maps, it is difficult to control the same dimension of channels shared by each scale feature map under the constraint of limited computation cost. Therefore, in the feature fusion of our neck, we adopt the setting of different scale feature maps with different dimensions of channels. Performance with the same and different channels as well as precision benefits from the Neck depth and width trade-offs are compared, Table.\ref{tab:featuremap_scale} shows the results. We can see that by flexibly controlling the number of channels in different scales, we can achieve much higher accuracy than sharing the same channels at all scales. Best performance is obtained when depth equals 3 and width equals (96, 192, 384).
2) GFPN enhances feature interaction by queen-fusion, but it also brings lots of extra upsampling and downsampling operators. The benefits of those upsampling and downsampling operators are compared and results are shown in Table.\ref{tab:queen_fusion}. We can see that the additional upsampling operator results in a latency increase of 0.6ms, while the accuracy improvement was only 0.3mAP, far less than the benefit of the additional downsampling operator. Therefore, under the constraints of real-time detection, we remove the extra upsampling operation in queen-fusion.
3) In the feature fusion block, we first replace original 3x3-convolution-based feature fusion with CSPNet and obtain 4.2 mAP gain. Afterward, we upgrade CSPNet by incorporating re-parameterization mechanism and connections of efficient layer aggregation networks (ELAN)~\cite{yolov7}. Without bringing extra huge computation burden, we achieve much higher precision. The results of comparison are listed in Table.\ref{tab:fusion}.
\begin{table}
    \begin{center}
    \caption{Ablation Study on the connection of queen-fusion. $\searrow$ and $\nearrow$ denote the upsampling and downsampling operations respectively.}
    \label{tab:queen_fusion}
    \setlength{\tabcolsep}{3pt}
    \begin{tabular}{cc|cccc}
    \toprule
    $\searrow$ & $\nearrow$ & Latency & FLOPs & AP   \\
    \midrule 
               &            & 3.62 & 33.3 & 44.2     \\
    \checkmark &            & 4.19 & 37.7 &             44.5    \\
               & \textbf{\checkmark} & \textbf{3.83} & \textbf{37.8} & \textbf{45.6}     \\
    \checkmark & \checkmark & 4.58 & 42.8 & 45.9     \\
    \bottomrule
    \end{tabular}
    \end{center}
\end{table}
\begin{table}
    \begin{center}
    \caption{Ablation study on the feature fusion style. CSP denotes the Cross-Stage-Partial Connection. Reparam \cite{ding2022re,ding2021repvgg} denotes applying re-parameter mechanism on the bottleneck of CSP. ELAN denotes the connections of efficient layer aggregation networks.}
    \label{tab:fusion}
    \setlength{\tabcolsep}{3pt}
    \begin{tabular}{l|ccccc}
    \toprule
    Merge$-$Style & Latency & FLOPs & AP    \\
    \midrule 
    Conv                      & 3.64 & 44.3 & 40.2  \\
    CSP                       & 3.72 & 36.7 & 44.4 \\
    CSP + Reparam            & 3.72 & 36.7 & 45.0 \\
    \textbf{CSP + Reparam + ELAN}  & \textbf{3.83} & \textbf{37.8} & \textbf{45.6} \\
    \bottomrule
    \end{tabular}
    \end{center}
\end{table}
\subsection{ZeroHead and AlignOTA}
In recent advancements of object detection, decoupled head is widely used~\cite{yolox,yolov6,yoloe}. With the decoupled head, those models achieve higher AP, while the latency grows significantly. To trade off the latency and the performance, we have conducted a series of experiments to balance the importance of neck and head, and the results are shown in Table.\ref{tab:neck_head_tradeoff}. From the experiments, we find that ``large neck, small head'' would lead to better performance. Hence, we discard the decoupled head in previous works~\cite{yolox,yolov6,yoloe}, but only left a task projection layer, \ie, one linear layer for classification and one linear layer for regression. We named our head as ZeroHead as there is no other training layers in our head. ZeroHead can save computations for the heavy RepGFPN neck to the greatest extent. It is worth noticing that ZeroHead essentially can be considered as a coupled head, which is quite a difference from the decoupled heads in other works~\cite{yolox,yolov5,yolov6,yoloe}. 
\begin{table}
    \begin{center}
    \caption{Studies on the balance between RepGFPN and ZeroHead. }
    \label{tab:neck_head_tradeoff}
    \setlength{\tabcolsep}{3pt}
    \begin{tabular}{lccccc}
    \toprule
     Neck(width/depth) & Head(width/depth) & Latency(ms) & AP    \\
    \midrule 
    \textbf{(1.0/1.0)} & \textbf{(1.0/0.0)} & \textbf{3.83} & \textbf{45.6} \\
    (1.0/0.50) & (1.0/1.0) & 3.79 & 44.9 \\
    (1.0/0.33) & (1.0/2.0) & 3.85 & 43.7 \\
    (1.0/0.0) & (1.0/3.0) & 3.87 & 41.2 \\
    \bottomrule
    \end{tabular}
    \end{center}
\end{table}
In the loss after head, following GFocal~\cite{li2020generalized}, we use Quality Focal Loss (QFL) for classification supervision, and Distribution Focal Loss (DFL) and GIOU loss for regression supervision. 
QFL encourages to learn a joint representation of classification and localization quality. DFL provides more informative and precise bounding box estimations by modeling their locations as General distributions. The training loss of the proposed DAMO-YOLO is formulated as:
\begin{equation}
     Loss = \alpha\;loss_{QFL} + \beta\;loss_{DFL} + \gamma\;loss_{GIOU}
\end{equation}
\begin{table}
    \begin{center}
    \caption{The comparison of different on MSCOCO val dataset. }
    \label{tab:label_assignment_comparison}
    \setlength{\tabcolsep}{3pt}
    \begin{tabular}{lccccc}
    \toprule
     Assigner & AP    \\
    \midrule 
    ATSS~\cite{zhu2020autoassign}  & 43.1    \\
    simOTA~\cite{ge2021ota}   & 44.2    \\
    TOOD~\cite{tood}  &  45.4      \\
    \textbf{AlignOTA}  &  \textbf{45.6}      \\
    \bottomrule
    \end{tabular}
    \end{center}
\end{table}
\begin{table}
    \begin{center}
    \caption{Studies on the distillation methods for DAMO-YOLO on MSCOCO val dataset. The baseline of student is 38.2.}
    \label{tab:distill}
 \setlength{\tabcolsep}{3pt}
    \begin{tabular}{lcc}
    \toprule
     Methods & Epochs & AP    \\
    \midrule 
    Mimicking \cite{li2017mimicking} & 36  & 40.2     \\
    MGD \cite{yang2022masked} & 36  & 39.6   \\
    CWD \cite{shu2021channel} & 36 &  \textbf{40.7}  \\
    \bottomrule
    \end{tabular}
    \end{center}
\end{table}
Besides head and loss, label assignment is a crucial component during detector training, which is responsible for assigning classification and regression targets to pre-defined anchors. 
Recently, dynamic label assignment such as OTA~\cite{ge2021ota} and TOOD~\cite{tood} is widely acclaimed and achieves significant improvements compares to static one~\cite{zhu2020autoassign}. 
Dynamic label assignment methods assign labels according to the assignment cost between prediction and ground truth, \eg, OTA~\cite{ge2021ota}. Although the alignment of classification and regression in loss is widely studied~\cite{tood,li2020generalized}, the alignment between classification and regression in label assignment is rarely mentioned in current works. 
The misalignment of classification and regression is a common issue in static assignment methods~\cite{zhu2020autoassign}. Though dynamic assignment alleviates the problem, it still exists due to the unbalance of classification and regression losses, \eg, CrossEntropy and IoU Loss~\cite{yu2016unitbox}. 
To solve this problem, 
we introduce the focal loss~\cite{lin2017focal} into the classification cost, and use the IoU of prediction and ground truth box as the soft label, which is formulated as follows:
\begin{equation}
\begin{split}
    AssignCost &= C_{reg} + C_{cls} \\
    \alpha &= IoU(reg_{gt}, reg_{pred}) \\
    C_{reg} &= - ln(\alpha) \\ 
    C_{cls} &= (\alpha - cls_{pred})^2 \times  CE(cls_{pred}, \alpha) 
\end{split}
\end{equation}
With this formulation, we are able to choose the classification and regression aligned samples for each target. Besides the aligned assignment cost, following OTA~\cite{ge2021ota}, we form the solution of aligned assignment cost from a global perspective.
We name our label assignment as AlignOTA. 
The comparison of label assignment methods is conducted in Table.\ref{tab:label_assignment_comparison}. We can see that AlignOTA outperforms all other label assignment methods. 
\begin{figure}[t]
    \centering
    \includegraphics[width=0.48 \textwidth]{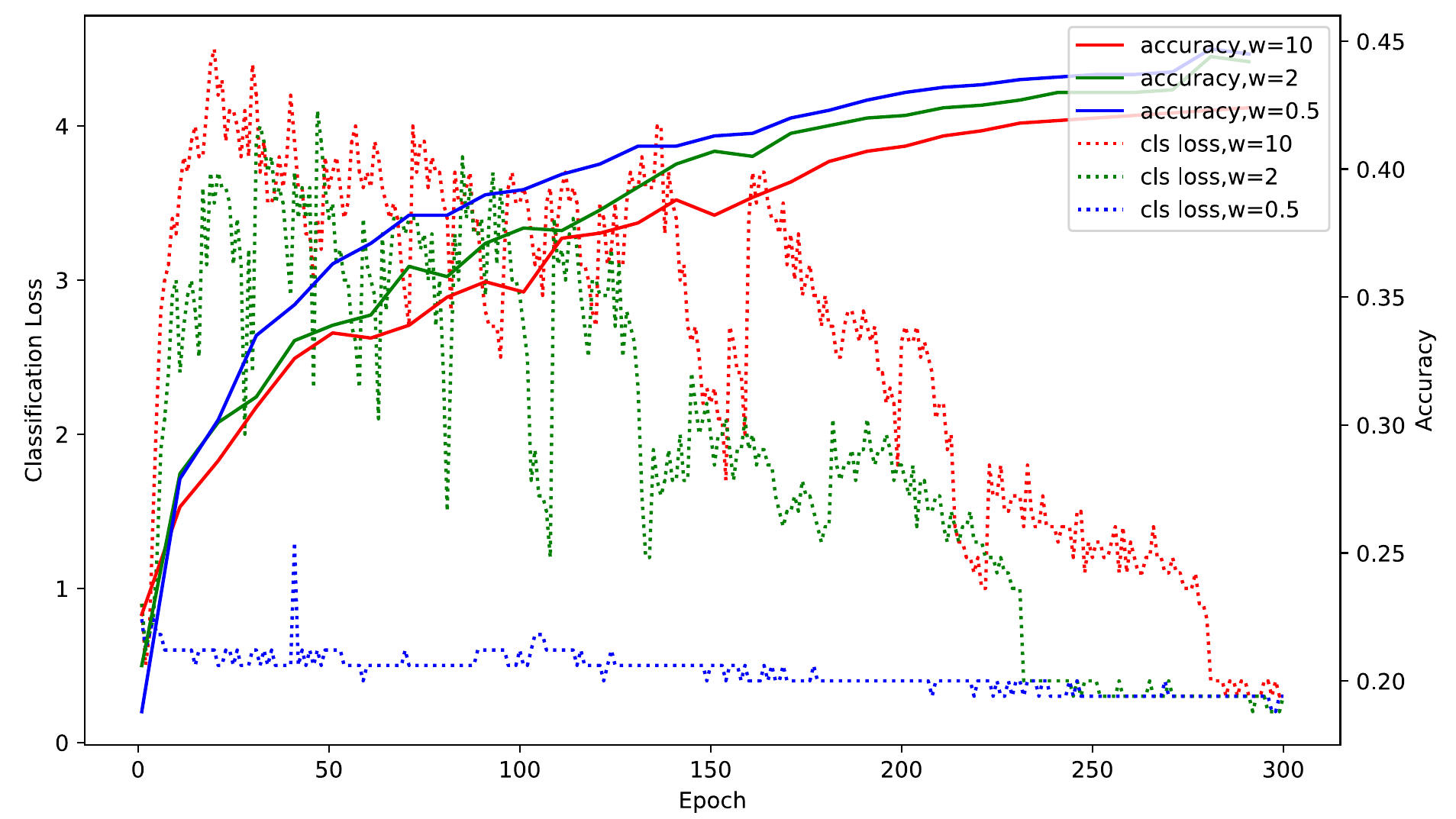}
    \caption{The classification loss and AP curves of distillation. The distillation loss weight is set to 0.5, 2, and 10 respectively. The classification loss has a significantly fast convergence with higher accuracy when the distillation loss weight is set to 0.5.}
    \label{fig:distill}
\end{figure}
\subsection{Distillation Enhancement}
Knowledge Distillation (KD) \cite{hinton2015distilling} is an effective method to further boost the performance of pocket-size models. Nevertheless, applying KD on YOLO series sometimes can not achieve significant improvements as hyperparameters are hard to optimize and features carry too much noise. In DAMO-YOLO, we first make distillation great again on models of all sizes, especially on the small size.
We adopt the feature-based distillation to transfer dark knowledge, which can distill both recognition and localization information in the intermediate feature maps \cite{huang2022masked}. We conduct fast validation experiments to choose a suitable distillation method for our DAMO-YOLO. The results are shown in Table.\ref{tab:distill}. We conclude that CWD is more fit for our models, while MGD is worse than Mimicking as complex hyperparameters make it not general enough.

Our proposed distillation strategy is split into two stages: 1) Our teacher distills the student at the first stage (284 epochs) on \textbf{strong mosaic} domain. Facing the challenging augmented data distribution, the student can further extract information smoothly under the teacher's guidance. 2) The student finetunes itself on \textbf{no mosaic} domain at the second stage (16 epochs). The reason why we do not adopt distillation at this stage is that, in such a short period, the teacher's experience will damage the student's performance when he wants to pull the student in a strange domain (\ie, no mosaic domain). A long-term distillation would weaken the damage but is expensive. So we choose a trade-off to make the student independent.

\begin{table*}[t!]
\begin{center}
\caption{Comparison with the state-of-the-art single-model detectors on MSCOCO validation set. * denotes using distillation. Latency is tested by ourself on T4 GPUs using TensorRT engine in FP16, while other results are from the corresponding papers. }
\label{coco_sota}
\setlength{\tabcolsep}{2pt}
\begin{tabular}{l|c|c|c|c|c|c|  c  c c c c}
\toprule
    Method  & Size & Latency(ms) & GFLOPs & Params(M) & FPS$^{V100}$ & AP & AP$^{50}$ & AP$^{75}$ & AP$^S$ & AP$^M$ & AP$^L$   \\
    \midrule
    YOLOX-T     & 416 & 1.78 & 6.5 & 5.1 & - & 32.8 & - & - & - & - & -    \\ 
    YOLOX-S     & 640 & 3.20 & 26.8 & 9.0 & 247 & 40.5 & - & - & - & - & -    \\ 
    YOLOX-M     & 640 & 6.46 & 73.8 & 25.3 & 177 & 46.9 & - & - & - & - & -    \\ 
    YOLOX-L     & 640 & 11.44 & 155.6 & 54.2 & 120 & 49.7 & - & - & - & - & -    \\ 
    \midrule
    YOLOv5-N     & 640 & 2.23 & 4.5 & 1.9 & - & 28.0 & 45.7 & - & - & - & -    \\ 
    YOLOv5-S     & 640 & 3.04 & 16.5 & 7.2 & 455 & 37.4 & 56.8 & - & - & - & -    \\ 
    YOLOv5-M     & 640 & 5.71 & 49.0 & 21.2 & 263 & 45.4 & 64.1 & - & - & - & -    \\ 
    YOLOv5-L     & 640 & 8.92 & 109.1 & 46.5 & 172 & 49.0 & 67.3 & - & - & - & -    \\ 
    \midrule
    YOLOv6-T     & 640 & 2.53 & 36.7 & 15.0 & - & 40.3 & 56.6 & - & - & - & -    \\ 
    YOLOv6-S     & 640 & 3.10 & 44.2 & 17.0 & - & 43.5 &60.4 & - & - & - & -    \\ 
    YOLOv6-M\textsuperscript{*}  & 640 & 5.72 & 82.2 & 34.3 & - & 49.5 & 66.8 & - & - & - & -    \\ 
    YOLOv6-L\textsuperscript{*} & 640 & 9.87 & 144.0 & 58.5 & - & 52.5 & 70.0 & - & - & - & -    \\ 
    \midrule
    YOLOv7-T-silu & 640 & 3.13 & 13.7 & 6.2 & - & 38.7 & 56.7 & 41.7 & 18.8 & 42.4 & 51.9    \\
    YOLOv7       & 640 & 9.08 & 104.7 & 36.9 & - & 51.2 & 69.7 & 55.9 & 31.8 & 55.5 & 65.0    \\
    \midrule 
    YOLOE-S      & 640 & 3.21 & 17.4 & 7.9 & 333 & 43.0 & 60.5 & 46.6 & 23.2 & 46.4 & 56.9    \\
    YOLOE-M      & 640 & 6.67 & 49.9 & 23.4 & 208 & 49.0 & 66.5 & 53.0 & 28.6 & 52.9 & 63.8    \\
    YOLOE-L      & 640 & 9.94 & 110.1 & 52.2 & 149 & 51.4 & 68.9 & 55.6 & 31.4 & 55.3 & 66.1    \\
    \midrule 
    DAMO-YOLO-T    & 640 & 2.78 & 18.1 & 8.5 & 397 & 42.0 & 58.0 & 45.2 & 23.0 & 46.1 & 58.5 \\
    DAMO-YOLO-T\textsuperscript{*} & 640 & 2.78 & 18.1 & 8.5 & 397 & 43.6 & 59.4 & 46.6 & 23.3 & 47.4 & 61.0   \\
    DAMO-YOLO-S    & 640 & 3.83 & 37.8 & 16.3 & 325 & 46.0 & 61.9 & 49.5 & 25.9 & 50.6 & 62.5   \\
    DAMO-YOLO-S\textsuperscript{*} & 640 & 3.83 & 37.8 & 16.3 & 325 & 47.7 & 63.5 & 51.1 & 26.9 & 51.7 & 64.9   \\
    DAMO-YOLO-M    & 640 & 5.62 & 61.8 & 28.2 & 233 & 49.2 & 65.5 & 53.0 & 29.7 & 53.1 & 66.1   \\
    DAMO-YOLO-M\textsuperscript{*} & 640 & 5.62 & 61.8 & 28.2 & 233 & 50.4 & 67.2 & 55.1 & 31.6 & 55.3 & 67.1  \\
    DAMO-YOLO-L & 640 & 7.95 & 97.3 & 42.1 & 126 & 50.8 & 67.5 & 55.5 & 33.2 & 55.7 & 66.6   \\
    DAMO-YOLO-L\textsuperscript{*} & 640 & 7.95 & 97.3 & 42.1 & 126  & 51.9 & 68.5 & 56.7 & 33.3 & 57.0 & 67.6  \\
    \bottomrule
    \end{tabular}
\end{center}
\end{table*}

\begin{table*}[t!]
\begin{center}
\caption{Comparison with the state-of-the-art light weight detectors on MSCOCO validation set. Latency is tested by ourself on Intel-8163 CPU with OpenVINO, while other results are from the corresponding papers.}
\label{coco_nano}
\setlength{\tabcolsep}{2pt}
\begin{tabular}{l|c|c|c|c|c|c|c|c|c|c}
\toprule
Method & Size & Latency(ms) & GFLOPs & Params(M) & AP & AP$^{50}$ & AP$^{75}$ & AP$^S$ & AP$^M$ & AP$^L$ \\
\midrule
Picodet-xs & 416 & 4.27 & 1.13 & 0.70 & 26.2 & - & - & - & - & - \\
Picodet-s & 416 & 4.53 & 1.65 & 1.18 & 32.5 & - & - & - & - & - \\
Picodet-m & 416 & 5.69 & 4.34 & 3.46 & 37.5 & - & - & - & - & - \\
Picodet-l & 416 & 7.03 & 7.10 & 5.80 & 39.4 & - & - & - & - & - \\
\midrule
Nanodet-plus-m & 416 & 5.80 & 1.52 & 1.17 & 30.4  & - & - & - & - \\
Nanodet-plus-m1.5x & 416 & 6.09 & 2.97 & 2.44 & 34.1 & - & - & - & - \\
\midrule
yolov8n & 640 & 6.16 & 8.7 & 3.2 & 37.3 & -- & - & - & - & - \\
\midrule
DAMO-YOLO-Ns & 416 & 4.08 & 1.56& 1.41  & 32.3 & 47.7 & 34.2 & 12.3 & 34.8 & 51.8 \\
DAMO-YOLO-Nm & 416 & 5.05 & 3.69& 2.14  & 38.2 & 54.7 & 40.8 & 20.2 & 41.7 & 57.6 \\
DAMO-YOLO-Nl & 416 & 6.69 & 6.04& 5.69  & 40.5 & 57.6 & 43.3 & 20.7 & 44.5 & 60.9 \\
\bottomrule
\end{tabular}
\end{center}
\end{table*}

In DAMO-YOLO, the distillation is equipped with two advanced enhancements: 1) Align Module. On the one hand, it is a linear projection layer to adapt student feature’s to the same resolution (${C, H, W}$) as teacher's. On the other hand, forcing the student to approximate teacher feature directly leads to minor gains compared to the adaptive imitation \cite{wang2019distilling}. 2) Channel-wise Dynamic Temperature. Inspired by PKD \cite{cao2022pkd}, we add a normalization to teacher and student features, to weaken the effect the difference of real values brings. After subtracting the mean, standard deviation of each channel would function as temperature coefficient in KL loss. 

Besides, we present two key observations for a better usage of distillation. One is the balance between distillation and task loss. As shown in Fig.\ref{fig:distill}, when we focus more on distillation (weight=10), the classification loss in student has a slow convergence, which results in a negative effect. The small loss weight\footnote{In our method, the cosine weight is utilized.} (weight=0.5) hence is necessary to strike a balance between distillation and classification. The other is the shallow head of detector. We found that the proper reduction of the head depth is beneficial to the feature distillation on neck. The reason is that when the gap between the final outputs and the distilled feature map is closer, distillation can have a better impact on decision.

In practice, we use DAMO-YOLO-S as teacher to distill DAMO-YOLO-T, and DAMO-YOLO-M as teacher to distill DAMO-YOLO-S, and DAMO-YOLO-L as teacher to distill DAMO-YOLO-M, while DAMO-YOLO-L is distilled by it self.

\begin{table}
    \begin{center}
    \caption{Ablation study on a large-scale dataset pre-trained with DAMO-YOLO-S. We validated our model on COCO without fine-tuning (denoted by '-'). For the VisDrone dataset, we loaded the pre-trained model and fine-tuned it for 300 epochs.}
    \label{tab:pretrain}
    \setlength{\tabcolsep}{3pt}
    \begin{tabular}{c|c|c}
    \toprule
    Pretrained Dataset & Downstream Task & AP    \\
    \midrule 
    COCO & - & 46.0  \\
    \textbf{COCO+Obj365+OpenImage} & \textbf{-} & \textbf{47.1} \\
    \midrule
    COCO & VisDrone & 24.6 \\
    \textbf{COCO+Obj365+OpenImage}  & \textbf{VisDrone} & \textbf{26.6} \\
    \bottomrule
    \end{tabular}
    \end{center}
\end{table}

\subsection{General Class DAMO-YOLO Model}
In this section, we introduce the General Class DAMO-YOLO model, which has been trained on multiple large-scale datasets, including COCO, Objects365, and OpenImage. Our model incorporates several improvements that enable high precision and generalization. Firstly, we address ambiguity resulting from overlapping categories across different datasets, such as "mouse," which can refer to a computer mouse in COCO/Objects365 and a rodent in OpenImage, by creating a unified label space for filtering. This reduces the original 945 categories (80 from COCO, 365 from Objects365, and 500 from OpenImage) to 701~\cite{zhou2021simple}. Secondly, we present a method of polynomial smoothing followed by weighted sampling to address the imbalance in dataset sizes, which often results in batch sampling biases towards larger datasets. Finally, we implement a class-aware sampling approach to tackle the issue of long-tail effects in Objects365 and OpenImage datasets. This assigns greater sampling weights to images containing categories with fewer data points.

By improving our model, we trained DAMO-YOLO-S on Large-Scale Datasets and achieved a 1.1\% mAP improvement compared to the DAMO-YOLO-S baseline on COCO. Furthermore, this pre-trained model can be fine-tuned for downstream tasks. We applied it to the VisDrone dataset for 300 epochs and achieved a 26.6\% mAP, surpassing the performance of the model pre-trained on COCO. These results highlight the advantages of large-scale dataset training and the robustness it provides to the model. The results are presented in Table.\ref{tab:pretrain}.

\section{Implementation Details}
Our models are trained 300 epochs with SGD optimizer. The weight decay and SGD momentum are 5e-4 and 0.9 respectively. The initial learning rate is 0.4 with a batch size of 256, and the learning rate decays according to a cosine schedule. Following YOLO-Series~\cite{yolox,yolov5,yolov6,yolov7} model exponential moving average (EMA) and grouped weight decay are used. To enhance data diversity, Mosaic~\cite{yolov4,yolov5} and Mixup~\cite{zhang2017mixup} augmentation is a common practice. However, 
recent advancement\cite{zoph2020learning,chen2021scale} shows that properly designed box-level augmentation is crucial in object detection. Inspired by this, we apply Mosaic and Mixup for image-level augmentation and employ the box-level augmentation of SADA~\cite{chen2021scale} after image-level augmentation for more robust augmentation.
\section{Comparison with the SOTA}

DAMO-YOLO release a series of general models and lightweight models, catering to both general scenarios and resource-limited edge scenarios. 

In order to evaluate the performance of DAMO-YOLO's general models against other state-of-the-art models, we conducted a comparative analysis on T4-GPU using the TensorRT-FP16 inference engine. The results, presented in Table.\ref{coco_sota}, demonstrate that DAMO-YOLO's general models outperform its rivals in terms of both accuracy and speed. Moreover, our proposed distillation technique offers significant improvements in accuracy. 

To assess the performance of light weight models, we conducted a comparative analysis on Intel-8163 CPU using Openvino as the inference engine. As shown in Table 2, DAMO-YOLO's lightweight model achieved substantial leading advantages, surpassing its competitors by a considerable margin in both speed and accuracy.

\section{Conclusion}
In this paper, we propose a new object detection method called DAMO-YOLO, the performance of which is superior to other methods in YOLO series. Its advantages come from new techs, including MAE-NAS backbone, efficient RepGFPN neck, ZeroHead, AlignedOTA label assignment and distillation enhancement.
{\small
\bibliographystyle{ieee_fullname}
\bibliography{damo_yolo}
}
\end{document}